\documentclass[acmsmall, nonacm]{acmart}
\AtBeginDocument{%
  }

\usepackage{amsmath,amsfonts}
\usepackage{algorithmic}
\usepackage{algorithm}
\usepackage{array}
\usepackage[caption=false,font=normalsize,labelfont=sf,textfont=sf]{subfig}
\usepackage{textcomp}
\usepackage{stfloats}
\usepackage{url}
\usepackage{cleveref}
\usepackage{verbatim}
\usepackage{graphicx}
\usepackage[nolist]{acronym} 
\usepackage{bm}
\renewcommand{\vec}[1]{\bm{#1}}

\usepackage{todonotes}
\usepackage{booktabs, multirow}

\begin{document}

\title{Performance and Explainability Requirements of Evolutionary Algorithms in Real-World Physics-Informed Optimization}
\renewcommand{\shorttitle}{Performance and Explainability Requirements of EAs in Real-World Physics-Informed Optimization}

 \author{Helena Stegherr}
 \email{helena.stegherr@uni-a.de}
 \orcid{0000-0001-7871-7309}
 \author{Michael Heider}
 \email{michael.heider@uni-a.de}
 \orcid{0000-0003-3140-1993}
 \author{Nils Meyer}
 \email{nils.meyer@uni-a.de}
 \orcid{0000-0001-6291-6741}
 \author{Tobias Thummerer}
 \email{tobias.thummerer@uni-a.de}
 \orcid{0000-0003-0908-0783}
 \author{Thomas Wendler}
 \email{thomas.wendler@med.uni-augsburg.de}
 \orcid{0000-0002-0915-0510}
 \author{Pierre Aublin}
 \email{pierre.aublin@uni-a.de}
 \orcid{0000-0002-8082-7811}
 \author{Ennio Idrobo-Àvila}
 \email{ennio.idrobo-avila@uni-a.de}
 \orcid{0000-0003-2014-0994}
 \author{Lars Mikelsons}
 \email{lars.mikelsons@uni-a.de}
 \orcid{0009-0005-9006-9726}
 \author{Sebastian Zaunseder}
 \email{sebastian.zaunseder@uni-a.de}
 \orcid{0000-0001-6114-3142}
 \author{Jörg Hähner}
 \email{joerg.haehner@uni-a.de}
 \orcid{0000-0003-0107-264X}
 \affiliation{%
   \institution{Universität Augsburg}
   \city{Augsburg}
   \country{Germany}
 }

\renewcommand{\shortauthors}{Stegherr et al.}

\begin{abstract}
Evolutionary computation offers a variety of tools to solve complex real-world optimization problems.
However, research often focuses on smaller, simplified problems and optimization algorithms that sometimes miss expectations in real-world scenarios.
Additionally, trust in the applied algorithm and the solutions it provides is often essential in such settings, but requires an understanding of the search process itself.
This leads to evolutionary computation often not being seriously considered by practitioners in many application contexts, among them physics-based modeling.
In this article, techniques from evolutionary computation are detailed that can alleviate these problems.
First, five real-world physics-based optimization problems are introduced and described by domain experts.
For each of these, the requirements for the evolutionary algorithm regarding performance and explainability to increase trust and usability are presented.
We found that all domain experts expect fast convergence to a good solution and want some explanations for how the results were formed, while other requirements strongly depend on the respective problem.
Finally, we present existing approaches that can be leveraged to improve those aspects of evolutionary algorithms but have to our knowledge never been employed in complex real-world scenarios.
This implies a gap between both domains that needs to be closed to exploit the full potential of evolutionary computation.
\end{abstract}


\maketitle

\section{Introduction}



Applying \ac{EC} techniques in real-world scenarios can come with a variety of decisions that should be made in close collaboration between domain experts of both fields.
This includes, among others, the description and formulation of the optimization problem, but also additional requirements regarding the efficiency and explainability of the optimizer~\cite{Osaba2021}.
However, in many cases the application of \ac{EC} methods to real-world problems is investigated by experts of only one of the two fields.
This can lead to either oversimplified problem formulations or the rejection of evolutionary approaches due to lack of knowledge in terms of how to configure those for the required performance and explain their process and results.
This paper therefore aims at detailing a suitable approach for utilizing \ac{EC} methods in real-world scenarios, specifically focusing on physics-informed \ac{EC} and \ac{SciML}.

Physics-informed \ac{EC} methods are versatile approaches to solve complex real-world problems where closed-form solutions are unavailable and the solution space cannot be explored within a feasible time or at a reasonable cost.
They are, for example, used in the reconfigurable intelligent surfaces codebook compilation~\cite{Papadopoulos2024}, but can also be combined with simulations to simheuristics that are able to provide robust solutions to setups that include stochastic elements~\cite{Chica2020}.
Additionally, \ac{ML} and \ac{EC} are two scientific sub-areas of AI that have already been examined in combination for interdisciplinary topics in various publications (cf.~\cite{Zhang:2011}).

Moreover, the intersection of physics-informed approaches, scientific computing, and \ac{ML} is referred to as \ac{SciML}, a domain where physical information and constraints are combined with large amounts of available data to generate more precise models \cite{Baker:2019}.
Here again, \ac{EC} methods prove to be valuable in many scenarios.
They can be effectively combined with physics-informed neural networks~\cite{Wong2025,Zhang2024}, for example, by performing neuroevolution~\cite{Sung2023}.

These developments likewise show an increased demand for explainability due to their commonly strong focus on application in complex scenarios.
Especially in real-world applications, practitioners are more likely to adopt a method if they can trust both its behavior and performance~\cite{Heider2026,Rodemann2025}.
This trust can be gained through detailed explanations of the methods and models, as well as their solutions and predictions.
In the \ac{SciML} domain, this need for explainability is voiced in~\cite{Rowan2025}.
One possible solution revolves around using an \ac{EC} approach to identify symbolic expressions for poorly explainable \ac{ML} models~\cite{Rackauckas:2021}, specifically utilizing genetic programming-based symbolic regression~\cite{Cao:2000, Raza:2012}.
This can increase the fidelity and interpretability of a physics-informed neural network~\cite{Murari2025}.
\ac{EC} approaches themselves, however, also require explanations and recent research is concerned with finding appropriate techniques~\cite{Bacardit2022,Zhou2024,Stein2025book}.


For investigating the application challenges of \ac{EC} in this context, we focus on providing detailed examples for the definition of five real-world optimization problems and the requirements stated by the domain experts towards the \ac{EC} approach.
This includes the necessary performance aspects of the evolutionary algorithm, as well as explanations for its search process and results.
We summarize and discuss these requirements, and based on this information, we examine existing applications of \ac{EC} to these or very similar problems to determine if and how any of the requirements are fulfilled.
Then, we highlight the techniques from the \ac{EC} domain that are already available to improve the performance and the explainability to meet the expectations of the problem experts.

With this, we provide valuable information for domain experts from both fields.
\ac{EC} practitioners can use these examples to assess common problem formulations and complexities, as well as the requirements and expectations towards the performance and explainability of the optimizer.
Problem experts, on the other hand, can use the details on existing approaches towards fulfilling the requirements to inform their decision when considering using an evolutionary algorithm.

The paper is structured as follows:
\Cref{sec:problem_descriptions} details the five physics-based optimization problems, \acl{SCARA}, \acl{FPP}, shape optimization with DeepONets, \acl{EIT}, and dynamic \acl{PET}.
It also provides all information on the requirements that are specified towards \ac{EC} methods in the respective scenario.
\Cref{sec:approaches} then summarizes these requirements, describes the current usage of evolutionary algorithms for the problems and details all additional existing and still missing strategies that are necessary to fulfill them.
Finally,~\Cref{sec:conclusion} concludes the paper and presents possible future work.

\section{Problem Statements and Descriptions}
\label{sec:problem_descriptions}

This section details five physics-informed real-world problems from various domains for which \ac{EC} methods may offer advantages for finding appropriate solutions.
For each of those, domain experts directly describe their requirements regarding the \ac{EC} method and its respective explainability.

\subsection{\acl{SCARA}}
\begin{figure}[!htpb]
    \centering
    \includegraphics[width=.5\linewidth]{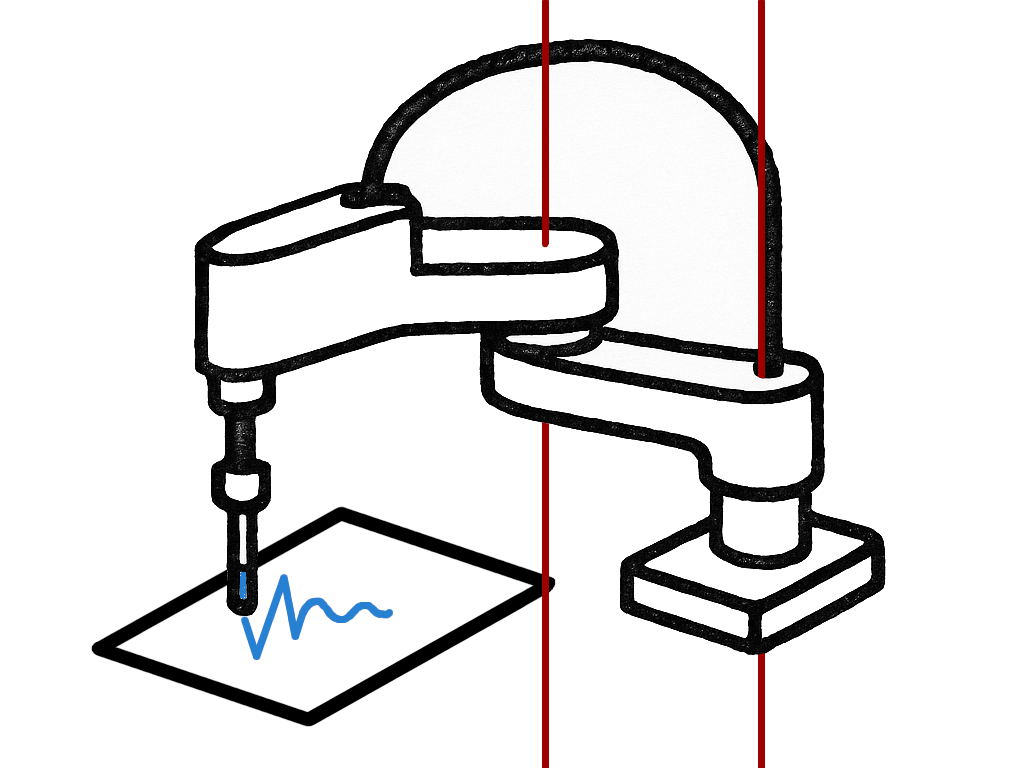}
    \caption{Illustration of the \ac{SCARA}, consisting of two actuated rotational axes (red), and a pen (blue) at the end-effector to draw on a sheet of paper.}
    \label{fig:SCARA}
\end{figure}

The goal of this example use case is to create a high-fidelity simulation model for a \acf{SCARA}, drawing with a pen on a piece of paper. A \ac{SCARA} typically consists of two rotational axes with the same orientation and possibly a third (often translational) axis, compare Fig. \ref{fig:SCARA}.

The friction force created by the contact between the pen and the paper must be overcome by the two motors that drive the two axes. The farther away the pen is from the mounting point of the robot, the more torque the motors have to apply to overcome static friction. As soon as static friction is overcome, the friction reduces abruptly and the generated torque must be reduced again to avoid overshooting. This characteristic effect is often referred to as \emph{stick-slip-friction} and is difficult to model and parameterize correctly due to the rapid (almost discontinuous) and highly nonlinear change in the friction coefficient. Nevertheless, accurate simulation models help in the development of robust control systems that can compensate for this effect (e.g., model-based feedforward control). In summary, modeling of the stick-slip effect is challenging but at the same time a very valuable measure. 

Generally speaking, this use case stands for the field of hybrid modeling \cite{Psichogios:1992, Schubert:1994, Sansana:2021}, i.e., simulation models are created based on physical equations and paired with ML models (e.g., \acp{ANN}) to compensate for model deviations that are typical for physical modeling---like the stick-slip example. 
    
As a starting point for hybrid modeling, a common simulation model is created based on physical equations. The simulation model consists of submodels of electric motors, robot axes (joints), and mechanical links. Without diving into details, modeling results in a right-hand side for the physics-based model
\begin{equation}
    \dot{\vec{x}}_\textrm{phy} = \vec{f}_\textrm{phy}(\vec{x}(t), \vec{\theta}_\textrm{phy}, t)
    ,
\end{equation}
with state derivative $\dot{\vec{x}}_\textrm{phy}$, state $\vec{x}$, physical parameters $\vec{\theta}_\textrm{phy}$ and time $t$. A detailed mathematical description of the model can be found at \cite{Thummerer:WS:2024b}.

However, the challenging stick-slip is explicitly not part of the modeling. This leads to a major deviation between the physics-based simulation model ("prediction") and the real world system ("data"), caused by neglecting the friction between pen and paper. The root of this deviation should be identified, and the cause should be compensated for, which can be done by incorporation of a feed-forward \ac{ANN}, resulting in a \emph{hybrid model}. For this use case, we modify the predicted dynamics of the physical model (the state derivative) by an \ac{ANN}, and obtain the right-hand side of the hybrid model $\vec{f}_\textrm{hyb}$ as 
\begin{equation}
    \label{eq:scara_righthand}
    \vec{f}_\textrm{hyb}(\vec{x}(t), \vec{\theta}_\textrm{ann}, \vec{\theta}_\textrm{phy}, t) = \vec{g}_\textrm{ann}(
        \vec{x}(t), \vec{f}_\textrm{phy}(
            \vec{x}(t), \vec{\theta}_\textrm{phy}, t
            ), 
        \vec{\theta}_\textrm{ann})
    ,
\end{equation}
where $\vec{g}_\textrm{ann}$ is the evaluation of the incorporated feedforward neural network, which calculates the hybrid model state derivative based on the system state, the state derivative of the physical model $\dot{\vec{x}}_\textrm{phy}$, and the parameterization $\vec{\theta}_\textrm{ann}$.

In summary, the core goal is to improve the prediction accuracy of the hybrid model compared to the physical model, which corresponds to a typical engineering scenario in the field.
The purpose of optimization (referred to as \emph{training} in the context of ML) is to identify the parameters of the \ac{ANN} so that a friction model for the end effector (pen) is learned, which imposes the corresponding resulting torques on the rotational axes of the \ac{SCARA}.
 
\begin{table}
    \caption{Variables for the \ac{SCARA} problem.}
    \label{tab:scara_variables}
    \begin{tabular}{lp{4.2cm}}
        \toprule
        $\alpha_i$ & Angle of axis $i$ \\
        $\dot{\alpha}_i$ & Angular velocity of axis $i$ \\
        $\ddot{\alpha}_i$ & Angular acceleration of axis $i$ \\
        $i_j$ & Current of motor $j$ \\
        $\vec{x}=\{ i_1, \alpha_1, \dot{\alpha}_1, i_2, \alpha_2, \dot{\alpha}_2 \}$ & System state\\
        $\vec{x}_0$ & Initial state \\
        $\tilde{\vec{x}}= \{ \alpha_1, \alpha_2 \} \subseteq \vec{x}$ & Measured system state\\
        $M = |\vec{x}| = 6$ & Count of states\\
        $\tilde{M} = |\tilde{\vec{x}}| = 2$ & Count of measured states\\
        $N$ & \hangindent=1em\hangafter=1 Number of points for the temporal discretization\\
        $\vec{X} \in \mathbb{R}^{N \times M}$ & ODE solution (states over time) \\
        $\vec{Y} \in \mathbb{R}^{N \times \tilde{M}}$ & Data (measured states over time)\\
        $\vec{f}_\textrm{phy} / \vec{f}_\textrm{hyb}$ & \hangindent=1em\hangafter=1 Right-hand side of the physical / hybrid model\\ 
        $\vec{g}_\textrm{ann}$ & Inference function of the \ac{ANN} \\
        $\vec{\theta}_\textrm{ann}$ / $\vec{\theta}_\textrm{phy}$ & \hangindent=1em\hangafter=1 Parameters of the \ac{ANN} / physical model\\
        \bottomrule
    \end{tabular}
\end{table}

\subsubsection{Problem statement}
Optimization of a hybrid model consists of two substeps:
First, the hybrid model is solved for a given initial state $\vec{x}_0$ and parameterization $\vec{\theta}$ to obtain the solution $\vec{X} \in \mathbb{R}^{N \times M}$ for the \ac{ODE} (cf.\@ Table~\ref{tab:scara_variables}). Note that the solution of the hybrid model therefore depends on the parameterization of the \ac{ANN} $\vec{\theta}_\textrm{ann}$, as well as the physical model $\vec{\theta}_\textrm{phy}$, so
\begin{equation}
    \label{eq:scara_pre_objective}
    \vec{X}(\vec{\theta}_\textrm{phy}, \vec{\theta}_\textrm{ann}) = \textrm{ODESolve}(\vec{f}_\textrm{hyb}, \vec{x}_0, t_0, t_f, \vec{\theta}_\textrm{phy}, \vec{\theta}_\textrm{ann})
    .
\end{equation}
The operator $\textrm{ODESolve}$ is an abstract placeholder for various \ac{ODE} solvers that solve a given right-hand side (here $\vec{f}_\textrm{hyb}$) starting with the initial state $\vec{x}_0$ from initial time $t_0$ to final time $t_f$, based on the given parameters $\vec{\theta}_\textrm{phy}$ (for the physical submodel) and $\vec{\theta}_\textrm{ann}$ (for the \ac{ANN} submodel). In addition, $\textrm{ODESolve}$ is a simplification of the actual interface and does not include, for example, event handling.

Second, since measurements in real systems can only be carried out for a finite number of measurement points, a temporal discretized loss function is used to compare the hybrid model solution $\vec{X}$ to real data $\vec{Y}$:
\begin{equation}
    \label{eq:scara_objective}
    \min_{\vec{\theta}_\textrm{ann}} \quad 
    \frac{1}{N} 
    \sum_{i=1}^{N} 
    \sum_{j=1}^{M} 
    \epsilon(\vec{X}(\vec{\theta}_\textrm{ann})_{i,j}, \vec{Y}_{i,j})
    ,
\end{equation}
where $\epsilon$ is an error function, usually the absolute or squared deviation is applied. As a final note, if the parameters of the physical model $\vec{\theta}_\textrm{phy}$ are uncertain, they can be optimized along the \ac{ANN} parameters. In general, this optimization problem is unconstrained.

\subsubsection{Problem analysis}
Concerning the optimization problem, typical sizes for the involved \acp{ANN} feature a parameter range between $50$ and $1,500$, depending on the \ac{ANN} layout (and therefore the requirements with respect to accuracy).
The optimization parameters and the system of equations are continuous. However, the model features time events to parse the table that holds the position of the end effector over time.
In general, there is no associated meaning for parameters within the \ac{ANN} (weights and biases). However, \emph{gates} can be introduced that allow a quantitative evaluation of the influence of signals (e.g., the dynamics) from the physical model and the \ac{ANN}~\cite{Thummerer:2022}.

The evaluation of the physical model \emph{before} the evaluation of the \ac{ANN} (which contains the parameters to be optimized) can be interpreted as a restriction of the state derivatives---and therefore, after numerical integration, also of the states. From this perspective, the introduction of the physical model could be interpreted as a hard constraint for the optimization problem. However, the constraint is structurally enforced and does not require explicit handling (in terms of choosing an optimizer that supports constraint optimization).

The general optimization problem is single-objective, however, additional regularization terms (\emph{explicit regularization}) can be added to the single objective.

The model is evaluated through solving an event-\acp{ODE} and comparing the solution with the ground truth data of the real system.
    
Efficient sensitivity analysis (gradient determination) for event-systems is an active topic of research. As it requires consecutive evaluations of the involved \ac{ANN}, differentiation through ODE solvers \cite{Sapienza:2024, Yingbo:2021} and event-handling, gradient determination for an \ac{ODE} solution is significantly more expensive than the evaluation of a feed-forward network. Inference, on the other hand, can be performed on highly optimized approaches. To summarize, the high computational cost for the gradient, while having a comparably low cost for model inference, together with the small parameter count (\emph{small} in the field of \ac{ML}), makes gradient-free optimization---especially \ac{EC}---a promising approach.  

\subsubsection{Requirements for evolutionary algorithms}
\paragraph{Optimization Process}
It is not required to find the global optimum. As ANNs are usually oversized w.r.t. the function to be learned, multiple acceptable solutions exist, and it is sufficient to find a ``good'' local optimum. In addition, finding a single solution is sufficient and having multiple solutions will in general lead to picking the best one (smallest loss).
Fast convergence is desirable because of the very expensive gradient determination, however, evolutionary algorithms have more room for achieving better results based on cheaper forward evaluations.
The optimization problem does not feature hard constraints in the sense of secondary conditions that must be fulfilled.
For the general case, no soft constraints need to be applied. However, soft constraints in terms of regularization can improve optimization convergence regarding various aspects and incorporation of soft constraints by extending the primary objective (adding additional terms) is very common in \ac{ML}.
The ANN inside the hybrid model is assumed to be a black-box, so no expert knowledge w.r.t. starting points for the optimization is available. However, common initialization routines for ANNs (e.g., \cite{Glorot:2010}) can be applied.

\paragraph{Explainability and Trust}
The core topic of interest is the found solution, and less the process to find it. 
Especially in deep neural networks research, a lack of explainability and trust is often tolerated and accepted. However, in case of hybrid modeling---featuring a significantly smaller parameter space compared to pure \ac{ML} models---information on how much of the search space the algorithm explored will allow developing trust in the solution found. If, e.g., the search space is well-explored, we trust that the local minimum found is a ``good'' one.
Robustness information and guarantees/uncertainty are not relevant for this specific use case.

\subsection{Fiber Patch Placement}
\acf{FPP} is a novel manufacturing technique for structures made from fiber-reinforced composite materials.
In this process, a robot places individual anisotropic composite patches such that the fibers align optimally with the load paths of a structural component. 
This results in tailored, variable-stiffness composites offering superior stiffness-to-weight ratios and reducing material waste from cutoffs \cite{Kussmaul.2019}. 
Optimization of such structures typically aims at finding positions and orientations \cite{Rettenwander.2014, Kussmaul.2019} or overlap patterns \cite{Fischer.2015} of the composite patches to improve the mechanical stiffness or strength of a structure. 

\begin{figure}[!htpb]
    \centering
    \includegraphics[width=.7\linewidth]{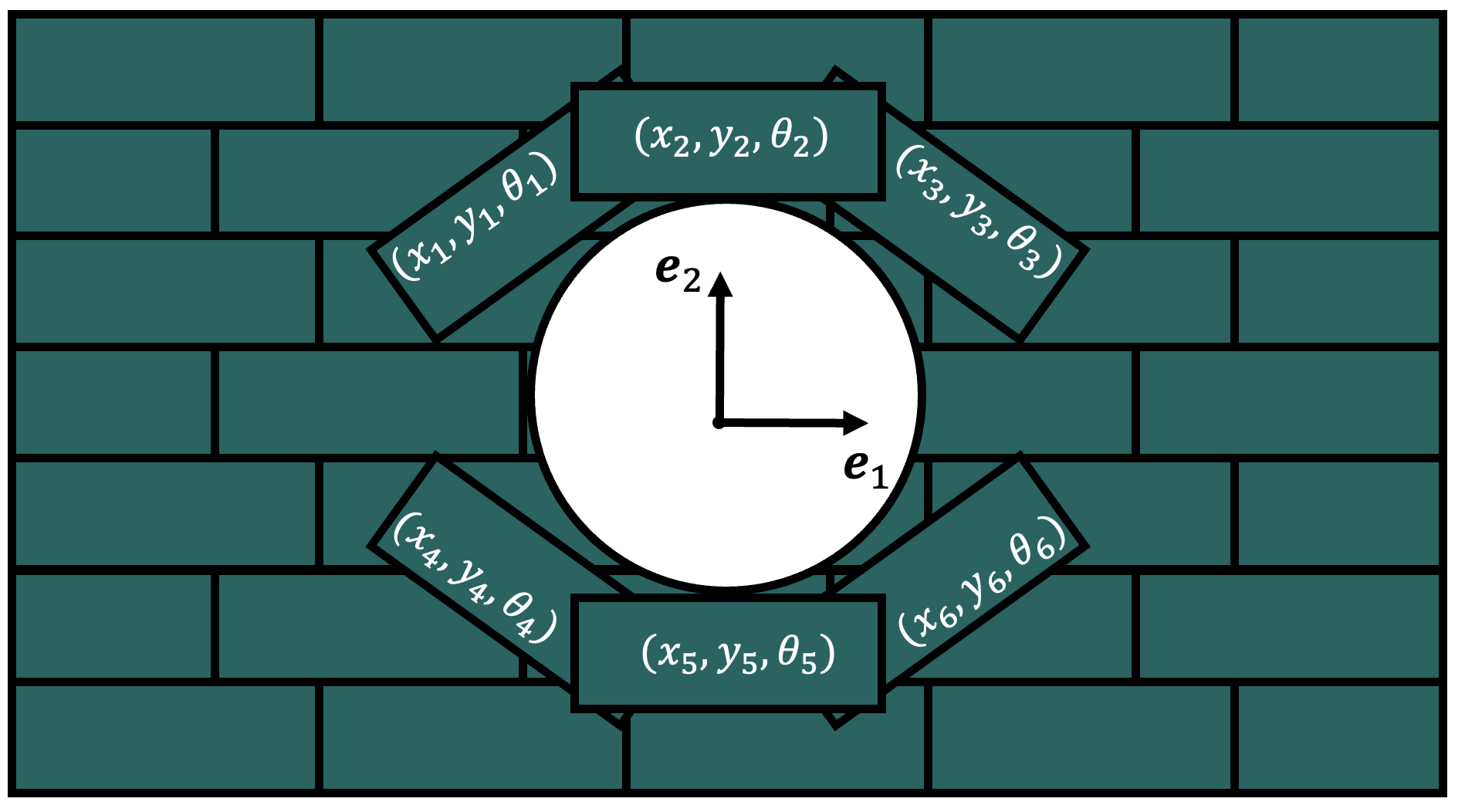}
    \caption{Illustration of a planar fiber patch placement problem: A structural component is assembled from $N$ anisotropic composite patches and we seek the optimal placement and orientation of those patches considering strength or stiffness.}
    \label{fig:fpp_example}
\end{figure}

\subsubsection{Problem statement}
Consider, for example, minimizing the compliance of a planar linear elastic hole plate (see Fig. \ref{fig:fpp_example} and Table~\ref{tab:fpp_variables}) with a strength constraint. 
The challenge is to find the position $(x_i, y_i)$ and orientation $\theta_i$ of each patch $i \in [1, N]$ to optimize
\begin{equation}
    \label{eq:fpp_objective}
    \min_{\{x_i, y_i, \theta_i\}_{i=1}^N} \quad \int_\Omega \pmb{\sigma}(\mathbf{x}) : \pmb{\varepsilon}(\mathbf{x}) \textrm{d}\Omega
\end{equation}
for positions $\mathbf{x}$ within a given spatial domain $\Omega$.
The colon operator ($:$) indicates the full inner contraction of two second rank tensors, i.e. $\pmb{\sigma} : \pmb{\varepsilon} = \sum_{i,j} \sigma_{ij} \varepsilon_{ij}$.
The optimization is subject to a strength constraint 
\begin{equation}
    \label{eq:fpp_strength_constraint}
    f\left(\pmb{\sigma}(\mathbf{x})\right) < 1 \quad \forall \mathbf{x} \in \Omega \\
\end{equation}
that evaluates a failure criterion $f$ and subject to a position constraint
\begin{equation}
    \label{eq:fpp_position_constraint}
    \Omega_i \subset \Omega
\end{equation}
ensuring that the patch domains $\Omega_i$ are entirely contained in the part domain $\Omega$.
Here, the stress tensor $\pmb{\sigma}(\mathbf{x})$ and infinitesimal strain tensor $\pmb{\varepsilon}(\mathbf{x})=1/2\left(\nabla \mathbf{u}(\mathbf{x}) + (\nabla \mathbf{u}(\mathbf{x}))^\top\right)$ are calculated from the displacement field $\mathbf{u}(\mathbf{x})$ as solution to the stationary elastic boundary value problem
\begin{align}
    \label{eq:fpp_momentum_balance}
    \quad \nabla \cdot \pmb{\sigma}(\mathbf{x}) &= \mathbf{0} &\forall \mathbf{x} &\in \Omega \\
    \label{eq:fpp_neumann}
    \pmb{\sigma}(\mathbf{x}) \cdot \mathbf{n}(\mathbf{x}) &= \mathbf{t}_0(\mathbf{x}) &\forall \mathbf{x} &\in \partial\Omega_\textrm{N}\\
    \label{eq:fpp_dirichlet}
    \mathbf{u}(\mathbf{x}) &= \mathbf{u}_0(\mathbf{x}) &\forall \mathbf{x} &\in \partial\Omega_\textrm{D} 
\end{align}
with an anisotropic linear material model 
\begin{equation}
    \label{eq:fpp_material_model}
    \pmb{\sigma}(\mathbf{x}) = \mathbb{C}(\mathbf{x}):\pmb{\varepsilon}(\mathbf{x}).
\end{equation}
The variable stiffness in the material model is computed as 
\begin{equation}
    \label{eq:fpp_variable_stiffness}
    \mathbb{C}(\mathbf{x}) = \sum_{\substack{i=1 \\ \mathbf{x} \in \Omega_i}}^{N} \frac{t_i}{T(\mathbf{x})}\mathbb{C}_i
\end{equation}
assuming equal strains for all layers in the laminate. 

\begin{table}
    \caption{Variables for the FPP problem.}
    \label{tab:fpp_variables}
    \begin{tabular}{lp{5.2cm}}
        \toprule
        $N$ & Number of patches \\
        $x_i, y_i$ & Center coordinates of patch i\\
        $\theta_i$ & Orientation angle of patch i\\
        $t_i$ & Thickness of patch i \\
        $\mathbb{C}_i \in \mathbb{R}^{2\times2\times2\times2}$ & Anisotropic stiffness of patch i \\
        $\Omega_i$ & Region covered by patch i \\
        $\mathbf{x} \in \Omega$ & Location in domain $\Omega \subset \mathbb{R}^2$\\
        $T(\mathbf{x})$ & Thickness field \\
        $\mathbf{u}(\mathbf{x}) \in \mathbb{R}^2 $ & Displacement field\\
        $\pmb{\sigma}(\mathbf{x}) \in \mathbb{R}^{2\times2} $ & Stress field\\
        $\pmb{\varepsilon}(\mathbf{x}) \in \mathbb{R}^{2\times2} $ & Strain field\\  
        $\mathbb{C}(\mathbf{x}) \in \mathbb{R}^{2\times2\times2\times2} $ & Stiffness field\\
        $f(\pmb{\sigma})$ & Failure criterion, for example Hashin \cite{Hashin.1980}\\
        $\mathbf{n} \in \mathbb{R}^2$ & Boundary normal on Neumann boundary $\partial \Omega_\textrm{N}$\\
        $\mathbf{t}_0 \in \mathbb{R}^2$ & Traction on Neumann boundary $\partial \Omega_\textrm{N}$\\
        $\mathbf{u}_0 \in \mathbb{R}^2 $ & \hangindent=1em\hangafter=1 Displacement on Dirichlet boundary $\partial \Omega_\textrm{D}$\\
        \bottomrule
    \end{tabular}
\end{table}

\subsubsection{Problem analysis}
The input design space is continuous and high-dimensional, as typical structures comprise $N=10^2-10^3$ fiber patches \cite{Kussmaul.2019}. 
Given a set of input tuples $\{x_i, y_i, \theta_i\}_{i=1}^N$, the resulting variable stiffness field across the domain can be computed using Equation \eqref{eq:fpp_variable_stiffness}.
For a prescribed stiffness distribution, the corresponding stationary elastic problem---defined by Equations \eqref{eq:fpp_momentum_balance} to \eqref{eq:fpp_material_model}---can then be solved for the displacement field $\mathbf{u}(\mathbf{x})$.  
This is achieved either with established numerical techniques, such as the finite element method, or with emerging approaches from scientific machine learning, such as physics-informed neural networks or MeshGraphNets \cite{Pfaff:2020}.
Solving this elastic problem is the computational bottleneck, particularly for large domains, thereby motivating the use of surrogate models to accelerate the evaluation.

In this case, the problem is formulated with a single objective in Equation \eqref{eq:fpp_objective}, with a strength constraint in Equation \eqref{eq:fpp_strength_constraint}, and a position constraint in Equation \eqref{eq:fpp_position_constraint} for a fixed number of patches. 
This will result in the stiffest structure that does not exceed a failure criterion and places all patches within the design domain. 
However, variants of this problem may be formulated, for example:
\begin{itemize}
    \item multi-objective optimization with trade-offs between the number of patches (=weight) and stiffness
    \item three-dimensional curved shell structures including kinematic draping effects
    \item incorporation of manufacturing uncertainties (limited position accuracy) and material uncertainties (properties vary between batches)
    \item non-linear material models such as damage models or non-linear geometries involving finite strains
\end{itemize}

The optimization landscape for this problem is non-convex and sensitivities are not readily available. 
This makes evolutionary algorithms particularly suitable for this problem.
        
\subsubsection{Requirements for evolutionary algorithms}
\paragraph{Optimization Process}
The primary objective is to identify the global optimum of the \ac{FPP} problem in a manner that is both robust and reproducible, i.e., the optimization process should reliably converge to the same solution for identical or similar problem parameters. 
Since evaluating the objective function and constraints requires solving a computationally expensive elastic problem, the optimization algorithm must achieve convergence in as few iterations as possible to accelerate the design process.

The optimization algorithm must strictly satisfy hard constraints, ensuring that patches are placed only within the design domain and that the resulting stress distribution does not lead to failure. 
Additionally, soft constraints may be introduced to reflect manufacturing constraints---for example, by penalizing large thickness jumps between adjacent patches to promote smooth transitions. 

A potential initialization strategy is to align the fiber patches according to the principal stress directions obtained from an isotropic, homogeneous elastic solution. 
However, as anisotropic patches are placed, they alter the stiffness distribution and thus the load paths within the structure, making it difficult to assess the quality of this initial guess a-priori.

\paragraph{Explainability and Trust}
Engineers have a certain intuition about where patches would need to be positioned.
If the solution is counterintuitive, users need a convincing explanation or interpretation of why patches end up at particular locations. 
Therefore, mechanistic explanations of the algorithm are desirable.
Ideally, the algorithmic configuration/parameters are also interpretable for easy tuning and should yield high reproducibility in the results. 

Additionally, it is desirable to provide information about the solution space to estimate how robust the solution is. 
A metric quantifying how much of the search space has been analyzed could help in tracking the progress and may support users in trading runtime vs. solution quality.

\subsection{Shape Optimization with DeepONets}
Shape optimization is a structural optimization type, in which we seek the optimal outer shape of a structure to enhance its performance, for example by reducing stresses to increase its lifetime.
Conventional shape optimization may be classified into spline-based methods and mesh-based methods. 
Spline-based methods result in CAD compatible results but require error-prone automatic re-meshing and a robust parametric model in the first place. 
Mesh-based methods parametrize the shape either with shape basis vectors, enabling rather complex differentiable shape modifications \cite{Biancolini.2020}, or bio-inspired growth rules \cite{Mattheck.1990}. 
However, mesh-based methods are prone to deterioration of element qualities. Two major drawbacks of current methods are the limited shape variety by parametrization and the need for a discrete mesh to solve the problem numerically.

In contrast to traditional mesh-based numerical methods, we want to employ mesh-less Deep Operator Nets \cite{Lu.2021} for solving the shape optimization problem.
These neural operators have been successfully employed to optimize airfoil shapes \cite{Shukla.2024} or inverse shape finding problems \cite{Zhang.2022}.

\subsubsection{Problem statement}
\begin{figure}[!htpb]
    \centering
    \includegraphics[width=.4\linewidth]{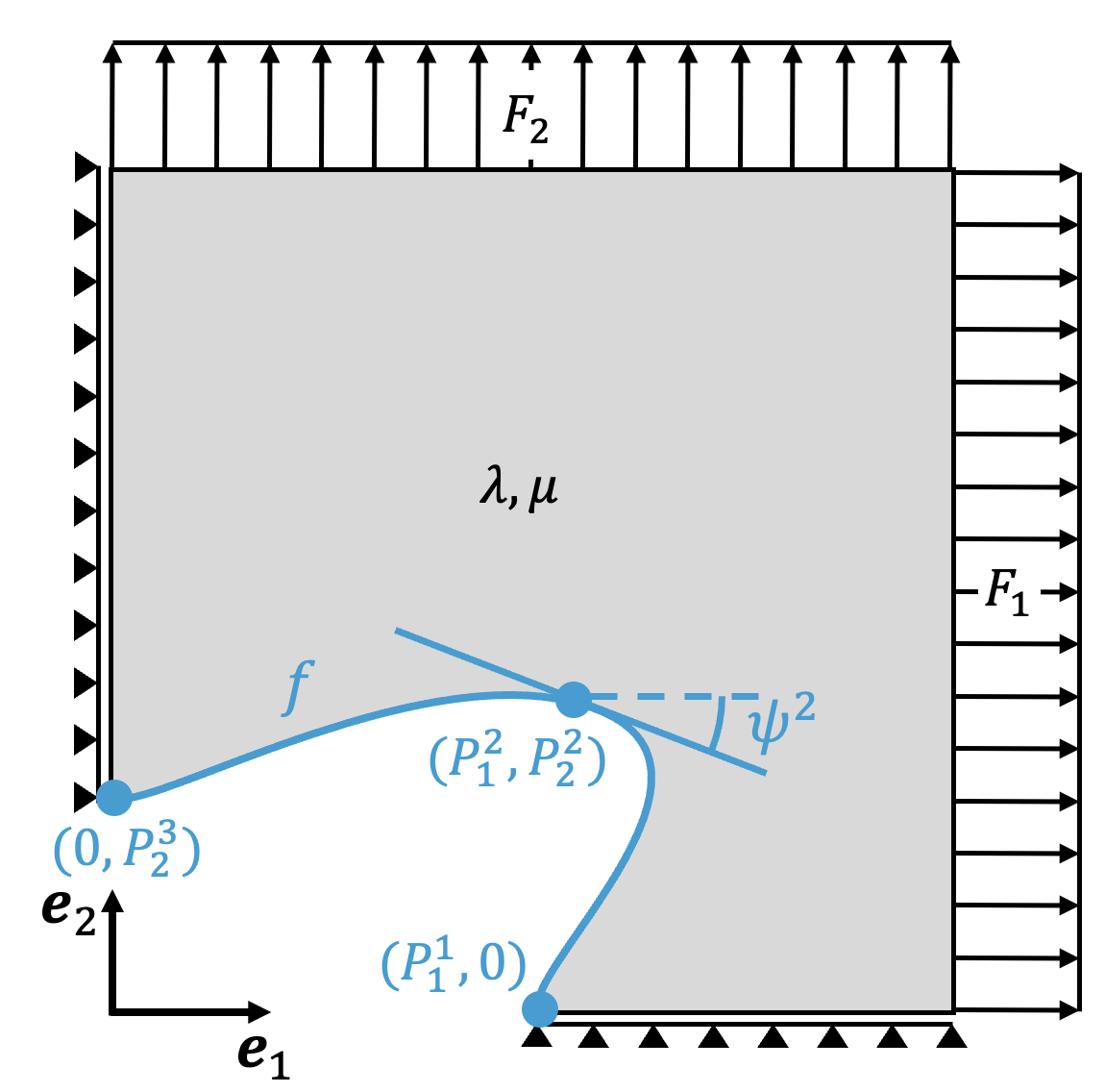}
    \caption{Illustration of the shape optimization problem: A quarter-symmetrical plate is loaded at two edges and we seek a hole shape that minimizes the maximum stress.}
    \label{fig:problem_shape_optimization}
\end{figure}

\begin{figure}[!htpb]
    \centering
    \includegraphics[width=.7\linewidth]{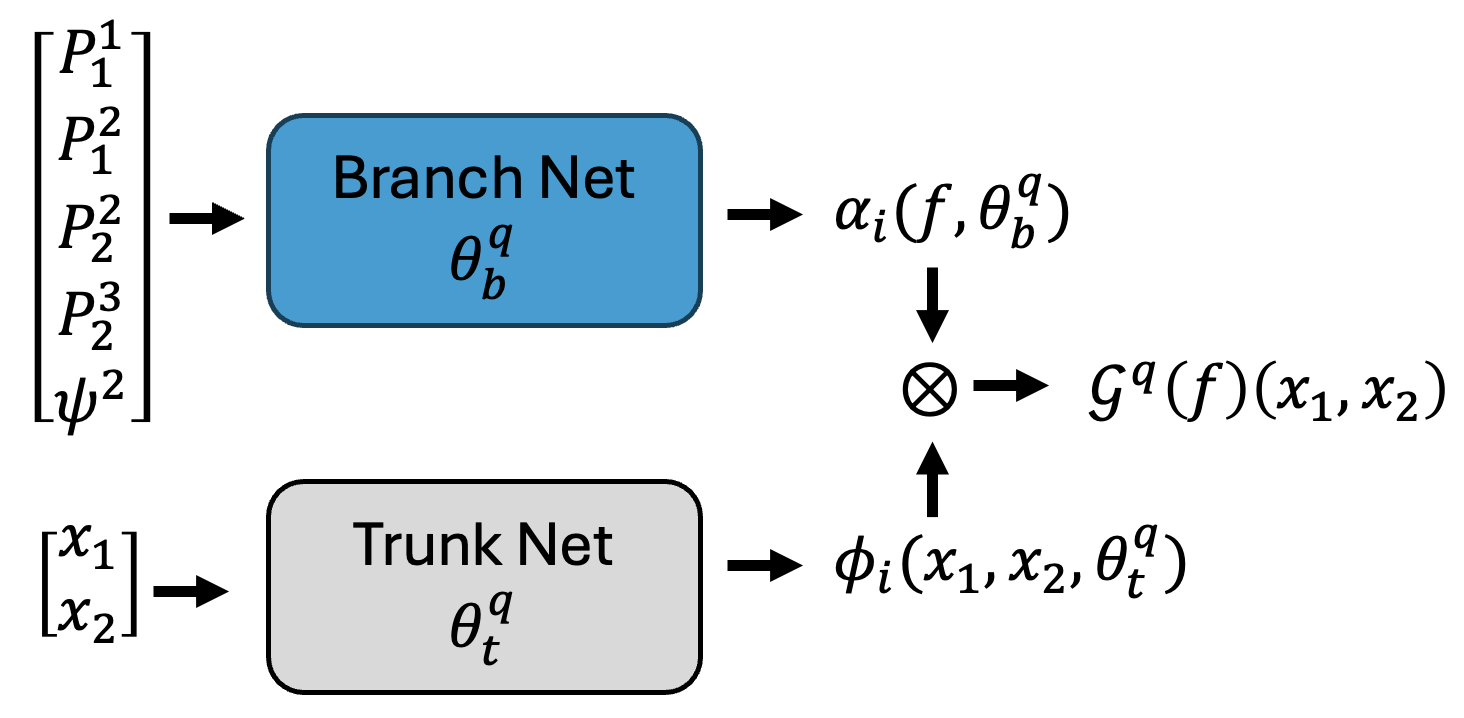}
    \caption{A DeepONet should learn an operator that maps an arbitrary input function $f$ describing the hole shape (embedded via a branch net) to a solution field for each position $(x_1, x_2)$, which is embedded via a trunk net.}
    \label{fig:problem_shape_deeponet}
\end{figure}

In this example, we consider a simple planar linear elastic plate with a parametric hole at its center (see Fig. \ref{fig:problem_shape_optimization} and Table~\ref{tab:shape_variables}) and want to identify the hole shape that minimizes the maximum of the norm of the deviatoric stress $\pmb{\sigma}'$ within the structure. The hole shape is described by a three-node B-Spline, which is parameterized with four modifiable node location parameters $P_d^n$ and an angle $\psi^n$, such that the optimization problem becomes
\begin{equation}
    \min_{\{P_1^1, P_1^2, P_2^2, P_2^3, \psi^2 \}} \max_{\mathbf{x} \in \Omega} ||\pmb{\sigma}'(\mathbf{x})||,
\end{equation}
where the deviatoric stress is computed from the same boundary value problem described in the previous example (Equations \eqref{eq:fpp_momentum_balance} to \eqref{eq:fpp_dirichlet}), but with an isotropic linear material model
\begin{equation}
    \pmb{\sigma} = \lambda \textrm{tr}(\pmb{\varepsilon}) \mathbf{I} + 2 \mu \pmb{\varepsilon}.
\end{equation}
To evaluate the boundary value problem efficiently for different curves, we seek an operator $\mathcal{G}$ that maps an input curve $f$ to output functions $\mathbf{u}(\mathbf{x})$ and $\pmb{\sigma}(\mathbf{x})$. 
This can be achieved with a DeepONet (see Fig. \ref{fig:problem_shape_deeponet}), in which a branch net computes an embedding of $f$ denoted $\alpha_i(f, \theta_b^q)$ and a trunk net computes an embedding of the domain $\mathbf{x}$ denoted $\phi_i(\mathbf{x}, \theta_t^q)$.
These embeddings are merged via a dot product operation to define the operator as 
\begin{equation}
    \mathcal{G}^q(f)(\mathbf{x}) = \sum_{i=1}^N \alpha_i(f,\theta_b^q) \phi_i(\mathbf{x}, \theta_t^q).
\end{equation}
Here, $q \in (u_1, u_2, \sigma_{11}, \sigma_{22}, \sigma_{12})$ indicates that different output function components are predicted with individual independent nets featuring parameters $\theta_b^q$ and $\theta_t^q$, respectively.

\begin{table}
    \caption{Variables for the shape optimization problem.}
    \label{tab:shape_variables}
    \begin{tabular}{ll}
        \toprule
        $P_d^n$ & B-Spline coordinate in direction $d$ for node $n$\\
        $\psi^n$ & B-Spline angle at node $n$\\
        $f$     & Function described by B-Spline \\
        $\lambda, \mu$ & Lamé parameters for the isotropic material \\
        $\theta_b^q$, $\theta_t^q$ & Parameters of the branch net and trunk net \\
        $\alpha_i$ & Input function embedding via branch net\\
        $\phi_i$ & Domain embedding via trunk net\\
        $\mathbf{x} \in \Omega$ & Location in domain $\Omega \subset \mathbb{R}^2$\\
        $\mathbf{u}(\mathbf{x}) \in \mathbb{R}^2 $ & Displacement field\\
        $\pmb{\sigma}(\mathbf{x}) \in \mathbb{R}^{2\times2} $ & Stress field\\
        $\pmb{\varepsilon}(\mathbf{x}) \in \mathbb{R}^{2\times2} $ & Strain field\\  
        $\mathbf{n} \in \mathbb{R}^2$ & Boundary normal on Neumann boundary $\partial \Omega_\textrm{N}$\\
        $\mathbf{t}_0 \in \mathbb{R}^2$ & Prescribed traction on Neumann boundary $\partial \Omega_\textrm{N}$\\
        $\mathbf{u}_0 \in \mathbb{R}^2 $ & Prescribed displacement on Dirichlet boundary $\partial \Omega_\textrm{D}$ \\
        \bottomrule
    \end{tabular}
\end{table}

\subsubsection{Problem analysis}
The number of design variables depends on how the input curve is parameterized. In this example, the curve is represented as a spline with five continuous design variables defining the positions and weights of its nodes. The DeepONet maps this input function to the corresponding static displacement and stress fields, a process that is computationally inexpensive during inference. Although this surrogate model does not suffer from data inaccuracies---being trained entirely on physics---its outputs still contain approximation errors. Therefore, it may be useful to occasionally solve the problem using other numerical techniques to obtain more accurate solutions, albeit at a higher computational cost.

The optimization problem is single-objective: minimizing the maximum norm of the deviatoric stress. Constraints can be introduced to avoid trivial solutions; for example, a lower bound on the area enclosed by $f$ can be imposed to prevent the hole from vanishing entirely.

Because the location of the maximum stress can shift abruptly, the optimization landscape is both non-convex and non-smooth. However, since DeepONet evaluations are relatively inexpensive, evolutionary algorithms are well-suited for exploring the design space efficiently.

\subsubsection{Requirements for evolutionary algorithms}
\paragraph{Optimization Process}
The main goal is to find the global optimum with fast convergence. Here, the algorithm should ideally make use of multi-fidelity outputs from the DeepONet and alternative mesh-based solution techniques.
It must account for hard constraints, such as ensuring that the spline is valid and bounds a fixed area. 
For the simple case of small-strain linear elasticity, there are analytical solutions to this specific problem \cite{Bjorkman.1976}. It may be used for validation or for initial parameters in variants with non-linear material.

\paragraph{Explainability and Trust}
The surrogate is associated with prediction uncertainty, hence potential optimal points should be checked with more expensive, but also higher fidelity FEM solutions to build trust in the optimization result. 
Therefore, treating uncertainty and multi-fidelity information is important for this type of problem. 

Additionally, it is desirable to provide information about the
solution space to estimate how robust the solution is. A metric
quantifying how much of the search space has been analyzed
could help in tracking the progress and may support users in
trading runtime vs. solution quality.

\subsection{Electrical Impedance Tomography}
\newsavebox{\myomega}
\savebox{\myomega}{$\scriptstyle\Omega$}

\begin{figure}
    \centering
    \includegraphics[width=.6\linewidth]{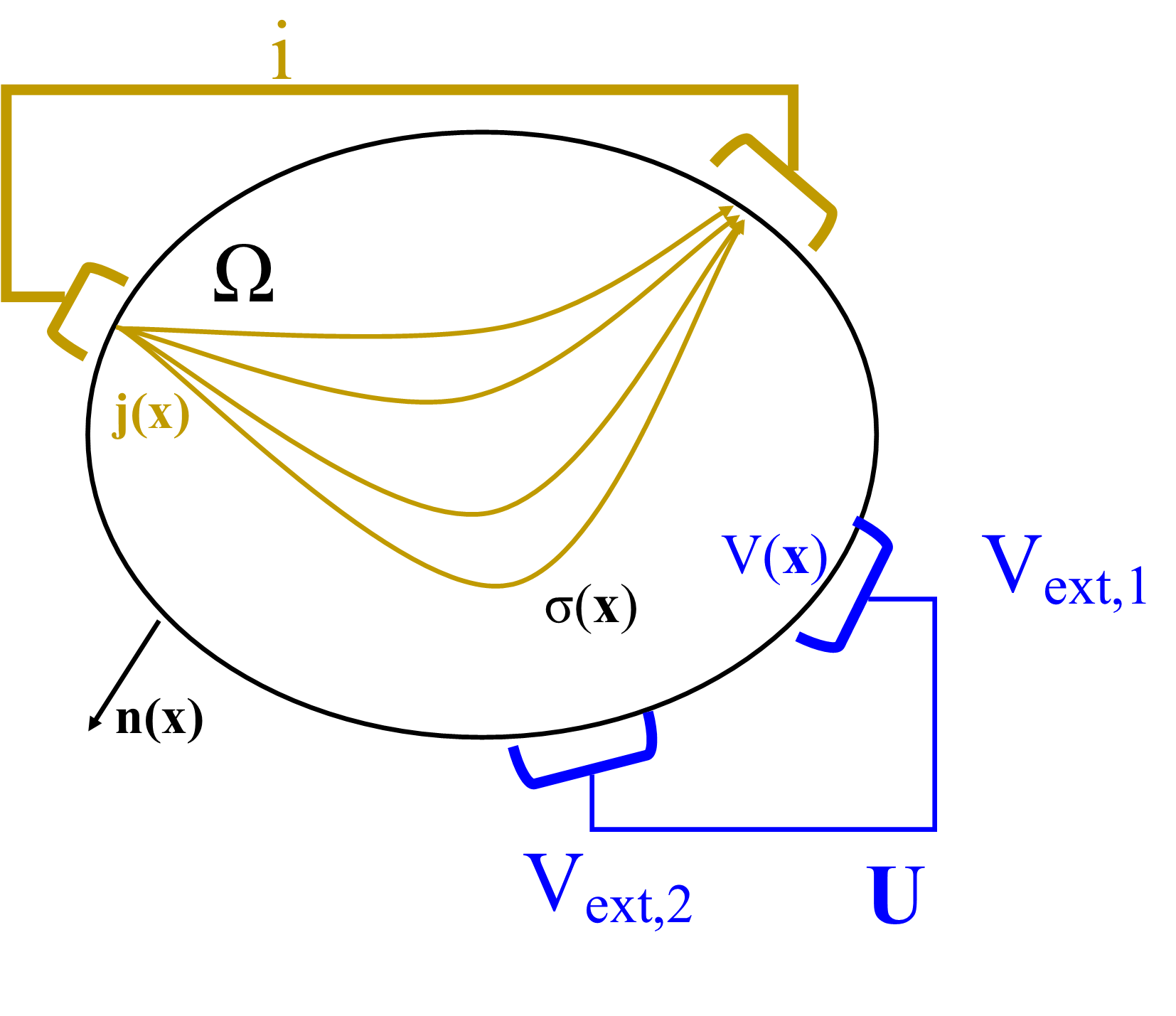}
    \caption{Illustration of the acquisition setup for \ac{EIT} and variables in the \ac{EIT} reconstruction problem. $\Omega$ is a volume in space which is applied a current $i$ at input electrodes. This results in a current density $\textbf{j}(\textbf{x})$ distribution inside $\Omega$ and a potential $V(\textbf{x})$ at position $\textbf{x}$ depending on the conductivity distribution $\sigma(\textbf{x})$. The potential matrix $\textbf{U}$ is obtained from external potential reading pairs $V_{ext,1}(\textbf{x})$ and $V_{ext,2}(\textbf{x})$ at boundary $\partial \Omega$ with the outward unit normal $\textbf{n}(\textbf{x})$ at position $\textbf{x}$.}
    \label{fig:eit-variables}
\end{figure}

Electrical conductivity is a property of materials describing their ability to transport electrical charges. This property varies among different biological tissues and can also be affected by physiological processes such as ventilation and perfusion. 
As illustrated in Fig. \ref{fig:eit-variables}, \ac{EIT} is a non-invasive imaging technique that reconstructs the internal electrical conductivity $\sigma$ distribution inside a volume $\Omega$ by injecting currents and measuring boundary voltages~\cite{DosSantos_deSouza_Ribeiro_Feitosa_Barbosa_Silva_Ribeiro_CovellodeFreitas_2018}. \ac{EIT} has diverse applications, ranging from medical imaging (e.g., lung function, gastric emptying, cardiac output estimations) to industrial process control. Its advantages include low cost, portability, and safety due to the absence of ionizing radiation. Nevertheless, the technique currently has limitations in spatial resolution.

\subsubsection{Problem Statement}
During the measurement, current densities $\textbf{j}$ and electrical potentials $V$ are subject to the partial differential equation
\begin{equation}
    \nabla(\sigma(\textbf{x})\nabla V(\textbf{x}))=0, \textbf{x}\in\Omega,
    \label{eq:eit_pde}
\end{equation}
with condition over the boundary $\partial \Omega$:
\begin{align}
    V(\textbf{x})=V_{ext}(\textbf{x}), \textbf{x}\in\partial\Omega \\
    \textbf{j}(\textbf{x})=-\sigma(\textbf{x})[\nabla (V(\textbf{x})) \cdot \textbf{n}(\textbf{x})], \textbf{x}\in\partial\Omega
\end{align}

where $\textbf{n}$ is the outward normal unit vector on the boundary surface (see also Table~\ref{tab:eit_variables}). 

\begin{table}
    \caption{Variables for the EIT problem modeling.}
    \label{tab:eit_variables}
    \begin{tabular}{lp{4.2cm}}
        \toprule
        $\partial$& Boundary\\
        $\nabla$& Nabla operator\\
        $\Omega$& Volume to image\\
        $\textbf{x}\in \Omega$& Position \\
        $\textbf{n}(\textbf{x})$& \hangindent=1em\hangafter=1 Normal outward unit vector at the boundary of $\Omega$\\
         $V(\textbf{x})$&Electrical potential at the position \textbf{x} \\
         $\textbf{U}$& Voltage matrix\\
         $\textbf{j}(\textbf{x})$& Current density at the position \textbf{x}\\
         $i=\int\int_S\textbf{j}(\textbf{x}) \textbf{n} \cdot dA(\textbf{x})\textbf{n}(\textbf{x})$& Current passing through a surface $S$ \\
         $\textbf{I}$& Current matrix\\
         $\sigma(\textbf{x})$ & Conductivity at the position \textbf{x} \\
         $\textbf{Y}(\sigma)$ & Admittance matrix \\
         \bottomrule
    \end{tabular}
    
\end{table}
As the measurements are driven by the injected current, the current density $\textbf{j}$ is known at the boundary. This and the measured potential differences at the electrodes provide a constraint $V_{ext}$ on the potential $V$ distribution inside the measured body. The variables are continuous.
A set $L$ of electrodes is placed around the zone to image. Usually, $L$ is even, and $L=8,16,32,64$ electrodes. At all electrodes, currents $I=(i_0...i_{L-1})^T$ are injected. Respectively, potential differences between electrode pairs are recorded at another set of electrodes ($V_{ext,1}(\textbf{x})$ and $V_{ext,2}(\textbf{x})$)~\cite{Youssef_Baby_Bedran_Doumit_El_Hassan_Maalouf_2024}. 
One hence obtains a vector $U$ of $R$ differences of potential between the electrodes, which leads to a linear equation in the form 
\begin{equation}
    \textbf{I}=\textbf{Y}(\sigma)\textbf{U}
\end{equation}
where $\textbf{Y}$ is an $L\times R$ matrix whose coefficients are dependent on the conductivity distribution inside the tissue $\sigma(\textbf{x})$.
The choice of the current injection and measurement strategies fixes the number of independent measurements available for the \ac{EIT} reconstruction as illustrated in Table \ref{tab:eit_measurements_indep_variables}.
\begin{table}
    \caption{Number of independent measurements obtained when using different current injection and voltage measurement strategies as described in 
    \cite{Youssef_Baby_Bedran_Doumit_El_Hassan_Maalouf_2024}.}
    \centering
    \begin{tabular}{c|c}
        \toprule
       Strategy  & Number of independent measurements \\
       \midrule
         Adjacent & $L(L-3)$ \\
         Cross method & $(L/2-1)(L-3)$\\
         Opposite method & $L(L-3)$\\
         Trigonometric & $(L/2)(L-1)$\\
         \bottomrule
    \end{tabular}
    
    \label{tab:eit_measurements_indep_variables}
\end{table}
Via finite element methods, the volume $\Omega$ is discretized and simulations based on known injected currents $i$ lead to electrical predicted potential and hence predicted voltages at the boundary $\hat{\textbf{U}}$.
The reconstruction algorithm has a single objective to optimize: it aims to find a distribution $\sigma$ within the body that minimizes the error $\epsilon(\sigma)=||\hat{\textbf{U}}-\textbf{U}||^2$.

\subsubsection{Problem Analysis}

Reconstruction algorithms are mostly assessed either through computational simulations or measurements based on phantoms for which conductivity distributions are known~\cite{Cheney_Isaacson_Newell_1999}. A set of quality criteria has been proposed to select good candidate solutions for reconstruction~\cite{Adler_Arnold_Bayford_Borsic_Brown_Dixon_Faes_Frerichs_Gagnon_Garber_2009}.

The problem is ill-posed as the minimizers of the reconstruction are not unique and the true conductivity distribution is not continuous (e.g., at the boundary between tissues).
As shown in Table \ref{tab:eit_measurements_indep_variables}, the problem is also ill-conditioned due to the sparsity of the available boundary measurements compared to the amount of finite elements in the measured volume, making it highly sensitive to noise and errors in the electrode positioning~\cite{Olmi_MBini_SPriori_2000}.
Current dominating reconstruction methods rely on the Newton-Raphson algorithm~\cite{Adler_Arnold_Bayford_Borsic_Brown_Dixon_Faes_Frerichs_Gagnon_Garber_2009}. 
To tackle the noise that could be introduced by the hardware, difference \ac{EIT} is often used. In that case, the difference of impedance between two sets of measurements is computed instead of absolute impedances, allowing for a linear approximation of the problem~\cite{Adler_Arnold_Bayford_Borsic_Brown_Dixon_Faes_Frerichs_Gagnon_Garber_2009}. However, the absolute \ac{EIT} aiming at recovering the impedance of the tissues remains harder to tackle.

Some applications focus on tissue characterization, and assume a static problem~\cite{dieter_haemmerich_vivo_2003}. However, there are cases where \ac{EIT} is used to observe dynamic processes such as in pulmonary function monitoring where ventilation changes the distribution of conductivity within the lungs~\cite{tomicic2019}.

\subsubsection{Requirements for evolutionary algorithms}
\paragraph{Optimization Process}

One would like to exclude solutions that are not consistent with the geometry of the volume or which propose conductivities of orders of magnitude that are not in the usual range of tissues, which would be around $10^{-1}$ S.m. or $1$ S.m. (unless expected, e.g., because of an implant).

Some other imaging modalities, like Magnetic Resonance Electrical Impedance Tomography (MREIT) can also provide a prior on the conductivity distribution~\cite{sajib_magnetic_2022} which can be used to set a constraint on the proposed solutions.

The \ac{EIT} inverse problem is characterized by its non-linear nature and often possesses a complex objective function landscape with multiple local minima~\cite{Zhou_2019}. \ac{EC} methods are inherently well-suited for tackling complex, non-linear, and ill-posed optimization problems, and are therefore full of potential, where traditional methods can easily fail \cite{Dimas_Alimisis_Uzunoglu_Sotiriadis_2024}. 
While traditional gradient-based optimization methods are prone to becoming trapped in these local optima and are highly sensitive to the initial values chosen for the reconstruction process, evolutionary algorithms are uniquely suited to navigate this complex, multi-modal optimization landscape, and to locate the true global minimum~\cite{Dimas_Alimisis_Uzunoglu_Sotiriadis_2024}. Their population-based approach allows for a broader exploration of the search space, significantly reducing the dependence on initial guesses and increasing the likelihood of identifying globally optimal solutions~\cite{Chen_Hu_He_Yang_Zhai_2010}. 
However, \ac{EC} algorithms may struggle with high dimensionality of the search space (e.g., 496 parameters for a typical FEM grid), which increases the complexity of finding optimal solutions~\cite{Kacarska_Loskovska_2002}. Furthermore, while \ac{EC} algorithms are known for their global search capabilities, they can sometimes require a greater number of iterations to converge to a desirable solution compared to more localized gradient-based methods~\cite{Zhou_2019}. This becomes problematic in applications such as lung and ventilation monitoring, where frame rates range from 25 to 50 images per second, each frame requiring reconstructions~\cite{tomicic2019}.

Beyond mere speed, the convergence must be stable. The reconstructed solution should not exhibit oscillations or divergence. Reconstruction algorithms applied to \ac{EIT} must demonstrate robustness against pervasive measurement noise and inherent modeling errors that characterize \ac{EIT} data acquisition, given the hardware limitations regarding the maximum number of electrodes and independent measurements. \ac{EIT} can therefore serve as a challenging testbed for \ac{EC} research to test new optimization algorithms~\cite{enriquez_philippine_2022}.

These opposite requirements encourage the development of hybrid \ac{EC} algorithms and advanced optimization methods to combine the expansive global search power of \ac{EC} with the faster local convergence characteristics of other methods~\cite{Dimas_Alimisis_Uzunoglu_Sotiriadis_2024}. For example, Particle Swarm Optimization has been effectively utilized for training \ac{ANN}, leading to both better and faster convergence in \ac{EIT} reconstruction~\cite{Martin_Choi_2016}.

Evolutionary algorithms, including Genetic Algorithms (GA), Particle Swarm Optimization (PSO), and Differential Evolution (DE), have been increasingly deployed in \ac{EIT} image reconstruction to tackle the complex, non-linear, and ill-posed inverse problem~\cite{Dimas_Alimisis_Uzunoglu_Sotiriadis_2024}. 
Some studies demonstrated that GA-EIT could outperform the Modified Newton-Raphson (MNR) algorithm in terms of stability, precision, and spatial resolution for static \ac{EIT} image reconstruction~\cite{Zhou_2019}, with more recent research exploring proportional GA to improve convergence rates and image quality~\cite{zhang_proportional_2020}. Historically, GA application in \ac{EIT} has been largely confined to two-dimensional experimental tank cases, indicating a need for further development for 3D and clinical scenarios~\cite{Dimas_Alimisis_Uzunoglu_Sotiriadis_2024}. PSO updates particle positions (candidate \ac{EIT} images) based on individual and global best-known positions \cite{Dimas_Alimisis_Uzunoglu_Sotiriadis_2024}, proving effective in overcoming Newton’s algorithm’s sensitivity to initial values~\cite{Mansouri_Alharbi_Haddad_Chabcoub_Alshrouf_Abd-Elghany_2021} and achieving faster, higher convergence rates compared to Back-Propagation (BP) algorithms for training ANNs for \ac{EIT} reconstruction \cite{Martin_Choi_2016}. DE is recognized for its simplicity and efficiency in optimization~\cite{Martínez-Guerrero_Lagos-Eulogio_Miranda-Romagnoli_Noriega-Papaqui_Seck-Tuoh-Mora_2024}, though it can be susceptible to premature convergence and stagnation in local minima~\cite{Martínez-Guerrero_Lagos-Eulogio_Miranda-Romagnoli_Noriega-Papaqui_Seck-Tuoh-Mora_2024}.
To address these, strategies such as modifying mutation, implementing self-adaptive parameter control, incorporating advanced population mechanisms, and hybridizing DE with other optimization algorithms have been developed.

\paragraph{Explainability}
As AI is increasingly integrated into healthcare, the need for trustworthy AI (TAI) systems has grown significantly \cite{Cappellini_Campagnola_Consales_2024}. TAI in healthcare is founded on core principles such as fairness, robustness, transparency, responsibility, privacy and security \cite{Cappellini_Campagnola_Consales_2024}. The lack of TAI systems hinders the transition of AI models from research to clinical practice, as the ``black-box'' nature of many AI models makes it difficult for healthcare professionals to interpret or trust their outputs, thereby affecting public acceptance \cite{Agafonov_Babic_Sousa_Alagaratnam_2024}. XAI aims to address the ``black-box'' problem inherent in complex AI models, making their decision-making processes interpretable for human users, especially clinicians \cite{Aziz_Manzoor_Mazhar_Qureshi_Qureshi_Rashwan_2024}. 
In the context of EIT, this involves understanding why a specific conductivity map was reconstructed or how the underlying algorithm arrived at a particular diagnosis or physiological assessment.

\subsection{Curve Fitting for Tracer Kinetic Modeling from dynamic Positron Emission Tomography sequences}

\subsubsection{Problem statement}
Dynamic \ac{PET} (dPET)---a 3D+t imaging modality---is used in clinical and research applications to analyze the dynamic uptake of positron-emitting tracers in the body. Tracers are radioactively labeled molecules such as sugar, neuroreceptors or antibodies, which enable imaging their dynamic distribution over time. Analyzing the dynamics of their uptake allows estimating relevant biomarkers such as vascularization, tracer inflow and outflow to or from cells, as well as metabolization or binding processes, all at millimeter-scale voxel resolution in 3D.

To analyze this data, so-called compartment models are employed. These models represent different compartments where the tracer may be located: for example, in the blood plasma or within tissue cells, either non-metabolized/free or metabolized/bound (which, depending on model complexity, can include multiple metabolized states)~\cite{carson_tracer_2005}. Most commonly used tracers can be modeled using 1- or 2-compartment models. An example is the irreversible 2-compartment model applied to 2-deoxy-2-[18F]fluoro-D-glucose (FDG), a glucose analogue. Such a model enables quantifying, in addition to vascularization, the active glucose transport into the cell, passive diffusion out of the cell and intracellular glucose metabolization~\cite{pantel2022principlesI}.

Historically, due to low spatial and temporal resolution, such analyses were performed only at the organ level---the so-called \emph{parametric analysis}~\cite{sokoloff1977}. Due to computational costs, this per-organ or per-lesion analysis remains the dominant clinical workflow. However, if each voxel is treated individually and the time-varying \ac{ODE}s from the compartment models are solved per voxel, it becomes possible to derive physiological parameters as continuous functions of space, resulting in fully 3D parametric images~\cite{kamasak_direct_2005}.

\begin{figure}[!htpb]
    \centering
    \includegraphics[width=.9\linewidth]{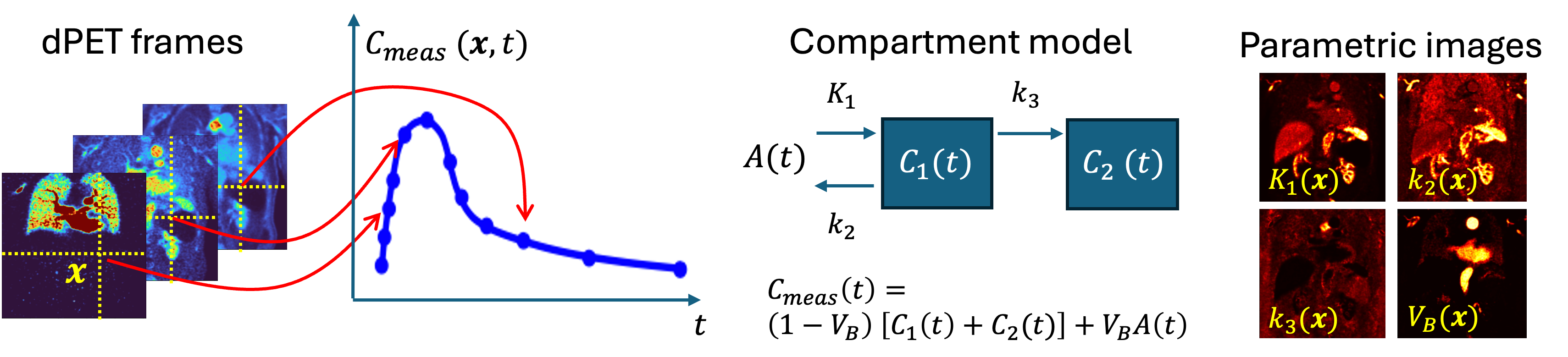}
    \caption{3D dynamic PET frames (dPET) depict the distribution of a radioactive tracer over space and time, and enable deriving time-activity curves (TACs), $C_{\text{meas}}(\textbf{x},t)$, for each voxel at position $\textbf{x}$. Typically, compartment models (in this figure, an irreversible 2-compartment model) assume the tracer can be in the blood supply (plasma) $A(t)$ or in multiple compartments $C_i(\textbf{x},t)$. By solving the resulting ODEs and performing curve fitting, parametric images are obtained, for example tracer inflow $K_1(\textbf{x})$, tracer outflow $k_2(\textbf{x})$, tracer metabolization $k_3(\textbf{x})$, and vascularization $V_B(\textbf{x})$.}
    \label{fig:pet_example}
\end{figure}

\subsubsection{Problem Analysis}
\paragraph{Mathematical formulation in the case of 1- or 2-Compartment Models}

\begin{table}
    \caption{Variables for the positron emission tomography problem.}
    \label{tab:pet_variables}
    \centering
    \begin{tabular}{lp{6cm}}
        \toprule
            $A(t)$ & input function (plasma tracer concentration, in Bq/ml) \\
            $C_1(\textbf{x},t)$ & \hangindent=1em\hangafter=1 tracer concentration in the first tissue compartment (free + non-specifically bound) \\
            $C_2(\textbf{x},t)$ & \hangindent=1em\hangafter=1 tracer concentration in the second tissue compartment (e.g., specifically bound or metabolized) \\
            $C_{\text{meas}}(\textbf{x},t)$ & measured total tracer concentration in the voxel \\
            $K_1(\textbf{x})$ & rate constant for plasma-to-$C_1$ transport \\
            $k_2(\textbf{x})$ & rate constant for $C_1$-to-plasma transport \\
            $k_3(\textbf{x})$ & rate constant for transfer from $C_1$ to $C_2$ \\
            $k_4(\textbf{x})$ & rate constant for transfer from $C_2$ back to $C_1$ \\
            $V_B(\textbf{x})$ & fractional blood volume in the voxel \\
        \bottomrule
    \end{tabular}
\end{table}


Dropping $\textbf{x}$ in the formulas in Table \ref{tab:pet_variables} for clarity, the system of \ac{ODE}s describing the compartment kinetics (see Fig.~\ref{fig:pet_example}) is:

\begin{equation}
    \frac{dC_1(t)}{dt} = K_1 A(t) - (k_2 + k_3) C_1(t) + k_4 C_2(t)
\end{equation}

\begin{equation}
    \frac{dC_2(t)}{dt} = k_3 C_1(t) - k_4 C_2(t)
\label{eq:C2}
\end{equation}

The measured \ac{PET} signal, also known as the time-activity curve (TAC), is given by:

\begin{equation}
    C_{\text{meas}}(t) = (1 - V_B) [C_1(t) + C_2(t)] + V_B A(t)
\label{eq:Cmeas}
\end{equation}

For a 1-compartment model, $k_3 = k_4 = 0$, equation~\ref{eq:C2} is not used, and $C_2(t)$ does not appear in Equation~\ref{eq:Cmeas}. For an irreversible 2-compartment model, as in the case of FDG, we set $k_4 = 0$.

\paragraph{Input and output}

The dPET frames, $C_{\text{meas}}(\textbf{x},t)$, are typically reconstructed from the list of detected coincidence events acquired by the \ac{PET} scanner during the interval of each frame. Due to the quantitative nature of \ac{PET}, the voxel intensities in the dPET frames represent the concentration of the radioactive tracer at each voxel in $\mathrm{Bq/ml}$ (decays per second per milliliter), which is a non-negative continuous value.

The number of dPET values depends on the image resolution in both space and time. For example, a full-body \ac{PET} scan at $2\,\mathrm{mm}$ isotropic resolution could yield volumes of size $512 \times 512 \times 800$ in spatial dimensions. Temporally, a high-sensitivity \ac{PET} device can produce frames as short as $2\,\mathrm{s}$, with total durations up to $120\,\mathrm{min}$. In practice, datasets typically contain 20–60 frames of non-uniform duration---short frames at the beginning to capture fast tracer dynamics, and longer frames toward the end to capture slower kinetics~\cite{sari_first_2022}.

Next to the dPET frames, the arterial input function $A(t)$ is required. This function can be obtained through direct arterial sampling, estimated from venous samples or extracted from dPET images themselves, typically by selecting a region of interest in a major artery---which explains the term ``arterial input''~\cite{volpi_update_2023}.

Depending on the tracer and model complexity, there can be multiple outputs, known as microparameters, which are non-negative continuous values. Simple one-compartment models include parameters such as tracer inflow $K_1(\textbf{x})$, tracer outflow $k_2(\textbf{x})$, and blood volume fraction $V_B(\textbf{x})$ per voxel. More complex two-compartment models add additional parameters, such as the rate of first-level tracer metabolization $k_3(\textbf{x})$ and the reconversion rate back to the original tracer $k_4(\textbf{x})$. These microparameters are not functions of time.

\paragraph{ODE solution and current curve fitting alternatives}

For 1- or 2-compartment models, an analytical solution to the ODE system exists. However, fitting the model parameters to measured data remains challenging because it requires solving a non-linear optimization problem. While basis function methods and iterative approaches using the Jacobian of the cost function---such as Levenberg-Marquardt optimization---are commonly employed, their performance is highly sensitive to the choice of basis function parameters or the initial values used in iterative solvers.

The main challenge arises from the \emph{ill-posedness} of the problem. dPET frames and input function values are extremely noisy, leading to a highly irregular optimization landscape with many local minima of similar cost. Measurement errors stem from the Poisson nature of radioactive decay, acquisition noise (scattering, energy resolution, timing resolution, electronic noise), image reconstruction artifacts and patient motion (respiratory, cardiac, voluntary)~\cite{miederer_machine_2023}. The arterial input function is also a significant source of error~\cite{volpi_update_2023}. Furthermore, since the model parameters appear as exponents in the analytical solution, small variations in data can cause large instabilities in parameter estimates~\cite{KANG202060}.

To improve stability, typical cost functions minimize the discrepancy between the measured TAC and the model prediction, commonly using $L_1$ or $L_2$ losses. However, these functions do not eliminate sensitivity to local minima.

In addition to robustness, efficiency is critical. Conventional per-voxel curve fitting approaches are computationally expensive. For example, fitting a single slice of size $400\times400$ across 60 frames can take approximately 20 minutes on a high-end personal computer (as of 2025). Processing a full-body scan of size $512 \times 512 \times 800 \times 60$ can take \emph{half a month}. While high-performance computing infrastructures reduce the time to several hours per patient, this remains prohibitive for large cohort studies, hyperparameter optimization or iterative workflows where parameter estimation is embedded as a sub-step. Additionally, hospital setups often lack such computational resources, making clinical deployment challenging.

To address these limitations, deep neural networks (DNNs) have been proposed as alternatives, both as patient-independent population-based solvers~\cite{de_benetti_self-supervised_2023} and patient-specific models~\cite{tehlan_physiological_2025}. While these approaches exploit spatial dependencies by considering neighboring voxels and significantly reduce computational cost, their numerical accuracy and stability are still inferior to conventional curve-fitting techniques.

\subsubsection{Requirements for evolutionary algorithms}

\paragraph{Optimization Process}
The ill-posedness and computational burden of voxel-wise parameter estimation make this problem a promising candidate for \ac{EC} approaches. The literature shows that evolutionary computation---especially GAs, PSO and related heuristics---has been successfully applied to curve fitting tasks in various scientific and engineering domains (e.g., ~\cite{GULSEN01071995,KARAKAPLAN2007235,svensson_using_2012,sabsch2017}). Nevertheless, their application to tracer kinetic modeling remains limited, with only two notable works: Huang et al.~\cite{huang2012} and Kang et al.~\cite{KANG202060}.

Huang et al.~\cite{huang2012} were, to the best of our knowledge, the first to apply PSO to this problem. They used PSO to estimate parameters in a complex three-compartment model for the amino acid tracer [18F]fluoro-3,4-dihydroxyphenyl-L-alanine in the brain---a tracer used to evaluate Parkinson’s disease. They demonstrated that, for such complex models, conventional curve fitting methods tend to overfit dominant parameters, leading to inaccurate estimates of less influential ones. PSO produced clinically plausible and numerically stable results for both healthy and Parkinsonian subjects.

In a more recent study, Kang et al.~\cite{KANG202060} explored PSO for solving the two-compartment curve-fitting problem. They proposed a modified loss function combining the conventional $L_2$ cost with an additional infinity-norm regularization term on the main exponents of the analytical solution, improving numerical stability. They compared PSO with proximal gradient descent and two PSO variants governed by uniform and Gamma-distributed random movements. The Gamma-based PSO yielded the best overall performance in terms of numerical accuracy, convergence speed and robustness, outperforming both Levenberg-Marquardt optimization and radial basis function methods, both in a simulation setup and in animal studies using FDG.

Building on this sparse but promising literature, we hypothesize that \ac{EC} methods could be further leveraged by incorporating spatial information across neighboring voxels, similar to recent deep learning approaches, while also embedding anatomical and physiological priors into the optimization process~\cite{miederer_machine_2023}. In such scenarios, the already noisy and multimodal optimization landscape becomes significantly more challenging, making conventional gradient-based methods prone to failure. EC-based strategies, which are better suited for rugged search spaces, may therefore provide more robust solutions and enable clinically meaningful parametric imaging.

\paragraph{Explainability}

Explanations of the search progress but especially the results are potentially helpful, depending on their presentation.
This is of increased importance when anatomical priors are used.
Additionally, the how much of the search space has been explored is of interest.
Information on robustness and uncertainties is also desirable but this may be unreliable due to data uncertainty.

\section{Approaches to Solve the Problems}
\label{sec:approaches}

We now take a closer look at the requirements for the EC methods that result from the five real-world problems described in~\Cref{sec:problem_descriptions}.
These explicitly include all requirements the respective experts deem essential for an application of EC methods in their domain.
We highlight common necessities regarding the EC method in general, as well as their explainability, and contrast this to the current application of those methods for similar problems.
Furthermore, we detail which techniques are already available to fulfill the requirements and where further research is required.

\subsection{Summary of Requirements}




In all five cases, an evolutionary optimization algorithm is a suitable choice.
However, the specific requirements towards the capabilities of the optimization algorithm vary between the use cases.
For the SCARA and PET problem, it is sufficient to find a good enough optimum that does not necessarily have to be the global optimum.
The FPP and EIT problems require robust and reproducible results, i.e.\@, the optimization algorithm should consistently result in very similar solutions and objective function values over multiple runs of the algorithm.
In case of the shape optimization use case, finding a solution very close to the global optimum is explicitly required.

Additionally, all five problems demand a fast converging optimization approach, where an adequate solution is provided after only few evaluations or little wall-clock time.
This is due to the real-world scenarios in which the optimization algorithms are applied, where the evaluation of the solutions often is very expensive, or a suitable solution has to be provided fast to continue with other operations.

Next, all five use cases require the incorporation of constraints for the solution variables, additionally to boundary constraints.
FPP, shape optimization, EIT and PET include hard constraints, i.e.\@, regions of the solution space that do not provide valid results and require specific handling.
Furthermore, SCARA, FPP and PET also need soft constraints to be considered during optimization.
This makes different approaches to constraint handling necessary.

Finally, for FPP, EIT, PET and shape optimization, additional information on solutions and respective objective values is available, e.g.\@, through literature or analytical solutions on smaller problem sizes.
These should be used to validate the optimizer, but the possibility to include this information as potential starting points for the optimization process is also required.

In addition to these requirements towards the capabilities of the optimization algorithm, all five use cases explicitly require explanations of the optimization approach.
In all cases, this results from a need for trust in the algorithm and its results.
Otherwise, classical but potentially less efficient approaches are preferred instead of evolutionary optimization algorithms.
More explicitly, for the FPP, shape optimization and EIT problems, the entire search process should be explained, i.e.\@, it is of interest how the optimization algorithm reached the final solution and what internal algorithmic dynamics contributed to this.
SCARA, FPP, EIT and PET require explanations of the results themselves.
This can include the influences of the individual solution variables on the results, but also the stability over several runs.
For both the search process and the final result, the influence of the algorithmic configuration, i.e.\@, the utilized components and parameter settings, necessitates further explanations, especially for the FPP, shape optimization and EIT problems.
In these cases, the practitioners want to know why a specific configuration is used and what contrasts other configurations.

Additionally, for FPP, shape optimization and PET, information on the robustness, quantification of uncertainties and guarantees for convergence or global optimality would be of interest, though explanations in this context are not considered strictly necessary.

Lastly, explanations with respect to the search space are considered.
For SCARA and PET, the search space coverage is of interest as this can indicate areas the optimization algorithm did not consider.
Surprisingly, in no case information gained through fitness landscape analysis was seen as relevant, even though this could provide more information on the underlying optimization problem.
Only for the shape optimization it was at least seen as possibly interesting.
After further discussion with the domain experts, this is likely due to little knowledge of research on fitness landscape analysis and the resulting information.
Providing details on fitness landscape analysis increased the interest of some domain experts, although they first requested some exemplary application to their problems to decide if this information could be useful.

\subsection{Available Techniques}






Before describing existing techniques to fulfill the requirements for the application of evolutionary optimization algorithms, we take a short look at how EC methods are currently applied to the five problems.
We thereby focus on use cases that are as similar as possible to the ones described in~\Cref{sec:problem_descriptions} and look if and how the stated requirements are approached.
This resulted in only few examples to be found because of the specificity of the use cases, but also due to a tendency of using simplified problems for EC research.

Similar to the SCARA problem as described in this work, the inverse kinematics problem in a SCARA has been tackled in~\cite{Bouzid2024}.
A neural network is used to solve the problem, while a genetic algorithm (GA) and particle swarm optimization (PSO) are compared in their capabilities to optimize the neural network parameters.
The overall goal of this study is to achieve an improved performance.
In this context, the algorithm consistency is investigated by examining and comparing the resulting error distributions.
This can be seen as some additional information on the robustness of the EC methods.
However, no other of the stated requirements has been considered.

For FPP and shape optimization, we could only find approaches that are strongly simplified or that have a different optimization goal.
In one instance, a GA is used in combination with a multifidelity Deep Operator Net, where the GA solves the inverse design problem but is not analyzed individually~\cite{Lu2022}.

The EIT problem shows the most applications of evolutionary optimization approaches, many of which are included in the problem description in~\Cref{sec:problem_descriptions}.
In most cases, these studies look for the evolutionary algorithm that results in the best performance, sometimes under consideration of the convergence speed in terms of function evaluations.
However, there is no dedicated effort towards the specified requirements, especially not regarding explainability.

Finally, approaches in dynamic PET are described in the aforementioned publications~\cite{huang2012,KANG202060}, as well as in~\cite{Li2018}, where a 3-Compartment model is optimized.
For this, a combination of an Artificial Immune System and a deterministic method is used.
The goal is to improve the global search capabilities in contrast to GA and PSO algorithms, thereby balancing the convergence speed.
This stands in contrasts to the requirements, as specifically fast convergence is needed but global optimality is not.
Huang et al.\cite{huang2012} and Kang et al.\cite{KANG202060} also focus on convergence, accuracy and in parts on robustness, however, other requirements are not discussed in any approach.

These examples indicate that many existing research on applying EC methods to the described optimization problems focuses on the performance, ignoring other important aspects of the optimization approach and its explainability that are necessary for real-world applications.
Within EC research, there are, however, many possibilities to incorporate the aforementioned requirements.







\subsubsection{Global optimization}
While in many cases, a good enough (local) optimum is sufficient, finding the global optimum is more difficult.
Generally, Evolution Strategies are deemed appropriate here, especially CMA-ES variants, as many improvement have been developed towards finding the global optimum of multimodal problems~\cite{Nobel2024}.
However, with evolutionary optimization algorithms, there is no guarantee that the global optimum is found in each run.
Explaining the results could be helpful to determine if a global optimum has been found or if a close local optimum might be good enough.

\subsubsection{Robustness and reproducibility} 
For evolutionary optimization algorithms, due to their stochastic search procedures, no guarantees can be given that they produce the same results in every run on the optimization problem.
State-of-the-art techniques are, however, often very robust on benchmark problems.
These can be used as a basis, while the robustness of their results can be examined through appropriate statistics and explanations.
Especially Bayesian statistics, which are increasingly common in benchmarking settings~\cite{Calvo2019,Mattos2021}, should be considered here due to their advantages in interpretability.

\subsubsection{Fast convergence}
Evolution Strategies and their variants can often be configured to converge fast.
However, not on all optimization problems, this leads to improved or even good results, and often there is a tradeoff between convergence speed and robustness~\cite{Glasmachers2025}.
Another possible approach would be to utilize evolutionary transfer learning methods, which have been shown to speed up convergence on expensive problems, though usually in a multi-objective optimization context~\cite{Liu2025,Chen2025}.

\subsubsection{Constraint handling}
An extensive overview on appropriate constraint handling techniques, including boundary, hard and soft constraints, is provided in~\cite{MezuraMontes2011}.
It is important to note that the most suitable technique depends on the algorithm as well as the optimization problem at hand.
    
\subsubsection{Including information}
In most cases, this aims at providing some initial solutions to the optimizer to either speed up the search process or facilitate finding a better final solution.
In most evolutionary optimization techniques, this is easily done during the initialization phase.
However, the initial solutions can strongly influence the search process~\cite{Sarhani2023}.
Including additional information has to be carefully evaluated to not bias the search in an unwanted direction.

\subsubsection{Explaining the search process and results}
The number of explainability techniques tailored to evolutionary optimization approaches is steadily growing (see, e.g.\@,~\cite{Stein2025book} for a recent overview).
Explanations for the search process can be provided through Search Trajectory Networks (STNs), which visualize the path of the algorithm through the search space~\cite{Ochoa2021}.
Furthermore, they can be used to compare the paths of different runs of the same algorithm, thereby displaying if their paths overlap.
Additionally, behavior analysis techniques are suitable to explain the search process, and even can provide details on the influence of the algorithmic components and parameter settings (e.g.~\cite{Nikolikj2025a}).
An example is the determination of algorithm-inherent biases towards specific regions of the search space~\cite{Vermetten2022}, or using unsupervised learning techniques to visualize similarities in behavior~\cite{Stegherr2025,Stegherr2025a,Nikolikj2025}.
\cite{Stein2025} also explored explanations and visualizations of component-influences in a benchmarking setting, providing strategies that can be transferred to real-world applications.

When aiming at explaining the results, search trajectories and sensitivity analysis of the solution variables has been explored, highlighting which variables contribute most to the final objective value~\cite{Fyvie2023,Fyvie2024}.
A similar approach examines the sensitivity of the results on the individual solution variables in real-world optimization problems from the energy domain~\cite{Lezama2023}.
The work also highlights the importance of such methods for increasing the understanding of the optimization algorithm and suggests other strategies, for example the aforementioned behavior analysis, as well as landscape analysis, to provide even more insights.
In addition, solution variable contributions have also been explored through the construction of a surrogate model~\cite{Singh2022} and in terms of robustness intervals, which depict ranges where a solution variable change is not reflected in a change of the objective value~\cite{Du2022}.

\subsubsection{Guarantees and uncertainty}
Providing guarantees for evolutionary algorithms' results and estimating the inherent uncertainty is difficult in many cases.
The common approach is therefore to provide an extensive statistical evaluation that indicates the respective qualities.
When the goal is to optimize a model under uncertainties using evolutionary techniques---as is often the case for problems of the SciML domain---several strategies are provided, e.g.\@, in~\cite{Ceberio2022} and~\cite{Bevia2022}.

\subsubsection{Search space and fitness landscape}
The search space coverage can be investigated by looking at STNs and similar approaches~\cite{Ochoa2021,Toda2022}.
Furthermore, there are visualizations for the entire population of solutions over the search process, also highlighting how much of the search space was visited during the algorithmic run~\cite{DeLorenzo2019}.

While details on the fitness landscape were not specifically requested for the five use cases, this information can still be used to supplement the other explanations, especially as algorithmic behavior and performance is highly problem-dependent.
The most relevant techniques in this area are exploratory landscape analysis and the resulting features that describe the fitness landscape~\cite{Malan2021,Renau2021}, local optima networks~\cite{Adair2019}, and extrema graphs~\cite{Sadler2023}.

\subsection{Research Gaps}




While there are numerous optimization strategies and explainability techniques available for evolutionary optimization, the main aspect that is missing is their application and evaluation in more complex real-world scenarios.
Especially as many approaches are problem-dependent, e.g.\@, the selection of an appropriate constraint handling technique or the best way to integrate prior information into the search process, exploring these for real-world optimization problems is necessary.
The same goes for landscape analysis techniques so that their capabilities are known to the respective problem domain experts and they can determine if the information gained through them is worthwhile.

In addition, which explanations are appropriate and useful for an expert on the problem, not the optimization algorithm, has to be explored~\cite{merry2021}.
For example, the details of the search process that are relevant and the way they are presented have to be determined in agreement with the recipients.
This might well be different for each use case, though a common procedure on how to determine the relevant information can be utilized (e.g.\@, similar to the assessments for evolutionary optimization in~\cite{Rodemann2025,Heider2026} or evolutionary machine learning in~\cite{Heider2023}).

\section{Conclusion}
\label{sec:conclusion}

This work explored the requirements for EC approaches when utilizing them to solve real-world physics-informed problems.
Five such real-world problems were detailed by the respective domain experts, who also stated their specific requirements regarding the EC algorithm's performance and its explainability.
For all cases, a fast converging approach was deemed necessary, though other optimizer requirements varied.
Furthermore, all domain experts demand some form of explainability, especially of the results, for the algorithm to be applied in the real-world scenarios.
When looking at existing research on EC methods, we found that many of those requirements could possibly be fulfilled through existing approaches, but these have not been explored much in real-world settings.

This should encourage researchers and practitioners from both the \ac{EC} and the respective problem domains to consider the underlying requirements and utilize the available techniques.
Through the presented real-world problems, \ac{EC} research has a foundation for what is necessary to get the developed methods into application.
However, this should not stop here but \ac{EC} experts are encouraged to engage with problem domain experts to gather relevant information on their expectations.
For problem domain experts, the summary on available \ac{EC} techniques serves as a guideline on what is possible with those methods, especially in terms of efficient problem solving and information that increases trust and understanding.

In future work, we would like to tackle one or more of the presented problems and include state-of-the-art EC techniques and explainability approaches to determine if and how this can satisfy the expectations of the domain experts.

\bibliographystyle{ACM-Reference-Format}
\bibliography{references.bib}

\begin{acronym}
    \acro{ANN}[ANN]{Artifical Neural Network}
    \acro{EC}[EC]{Evolutionary Computation}
    \acro{EIT}[EIT]{Electrical Impedance Tomography}
    \acro{FPP}[FPP]{Fiber patch placement} 
    \acro{ML}[ML]{Machine Learning}
    \acro{ODE}[ODE]{Ordinary Differential Equation}
    \acro{PET}[PET]{Positron Emission Tomography}
    \acro{SCARA}[SCARA]{Selective Compliance Assembly Robot Arm}
    \acro{SciCo}[SciCo]{Scientific Computing}
    \acro{SciML}[SciML]{Scientific Machine Learning}
\end{acronym}

\end{document}